\documentclass[10pt,twocolumn,letterpaper]{article}

\usepackage[pagenumbers]{cvpr} 

\definecolor{cvprblue}{rgb}{0.21,0.49,0.74}
\usepackage{kotex}
\usepackage{booktabs}
\usepackage{multirow}
\usepackage[normalem]{ulem}
\useunder{\uline}{\ul}{}
\usepackage{algorithm}
\usepackage{algorithmic}
\usepackage{verbatim}
\usepackage{makecell}
\usepackage{subcaption} 
\usepackage{caption}
\usepackage{adjustbox}
\usepackage{pifont}
\usepackage[pagebackref,breaklinks,colorlinks,allcolors=cvprblue]{hyperref}

\def\paperID{13093} 
\def\confName{CVPR}
\def\confYear{2025}

\title{HOT: Hadamard-based Optimized Training}

\author{
    Seonggon Kim\textsuperscript{1}, 
    Juncheol Shin\textsuperscript{2}, 
    Seung-taek Woo\textsuperscript{2}, 
    and Eunhyeok Park\textsuperscript{1,2}\\
    \\
    \textsuperscript{1}Department of Computer Science and Engineering, POSTECH\\
    \textsuperscript{2}Graduate School of Artificial Intelligence, POSTECH\\
    {\tt\small \{sungonuni, jchshin, wst9909, eh.park\}@postech.ac.kr}
}

\begin{document}
\maketitle
\begin{abstract}
It has become increasingly important to optimize backpropagation to reduce memory usage and computational overhead. Achieving this goal is highly challenging, as multiple objectives must be considered jointly while maintaining training quality. In this paper, we focus on matrix multiplication, which accounts for the largest portion of training costs, and analyze its backpropagation in detail to identify lightweight techniques that offer the best benefits. Based on this analysis, we introduce a novel method, Hadamard-based Optimized Training (HOT). In this approach, we apply Hadamard-based optimizations, such as Hadamard quantization and Hadamard low-rank approximation, selectively and with awareness of the suitability of each optimization for different backward paths. Additionally, we introduce two enhancements: activation buffer compression and layer-wise quantizer selection. Our extensive analysis shows that HOT achieves up to 75\% memory savings and a 2.6$\times$ acceleration on real GPUs, with negligible accuracy loss compared to FP32 precision. Our code is available at \url{https://github.com/sungonuni/HOT}.
\end{abstract}    
\section{Introduction}
\label{sec:intro}

Recent advances in combining large-scale foundation models with transfer learning have enabled superior performance on new tasks with limited data. However, with the rise of resource-intensive tasks such as high-resolution image generation~\cite{rombach2022high, zhang2022styleswin, shi2024resmaster}, video synthesis~\cite{voleti2022mcvd, singer2022make, blattmann2023align}, and long-context language processing~\cite{chen2023extending, ding2023longnet, li2024long, fang2024unimem}, fine-tuning now requires even more resources, making it nearly unaffordable for most companies or academic institutions.

However, designing an efficient training pipeline is highly challenging, as it involves numerous considerations and multiple tensors with diverse characteristics. Training optimization faces three key challenges: (1) memory usage for model parameters and optimizer states, (2) memory usage for intermediate activations needed for backpropagation (BP), and (3) accelerating BP. While the first challenge is well-addressed by recent advances like Parameter-efficient Fine-tuning (PEFT), such as LoRA~\cite{lora}, the other two challenges remain unsatisfactorily resolved. Approaches like LUQ~\cite{luq} and LBP-WHT~\cite{lbp-wht} focus on BP acceleration but fail to tackle all these objectives jointly.

\begin{figure}
    \centering
    \includegraphics[width=1.0\linewidth]{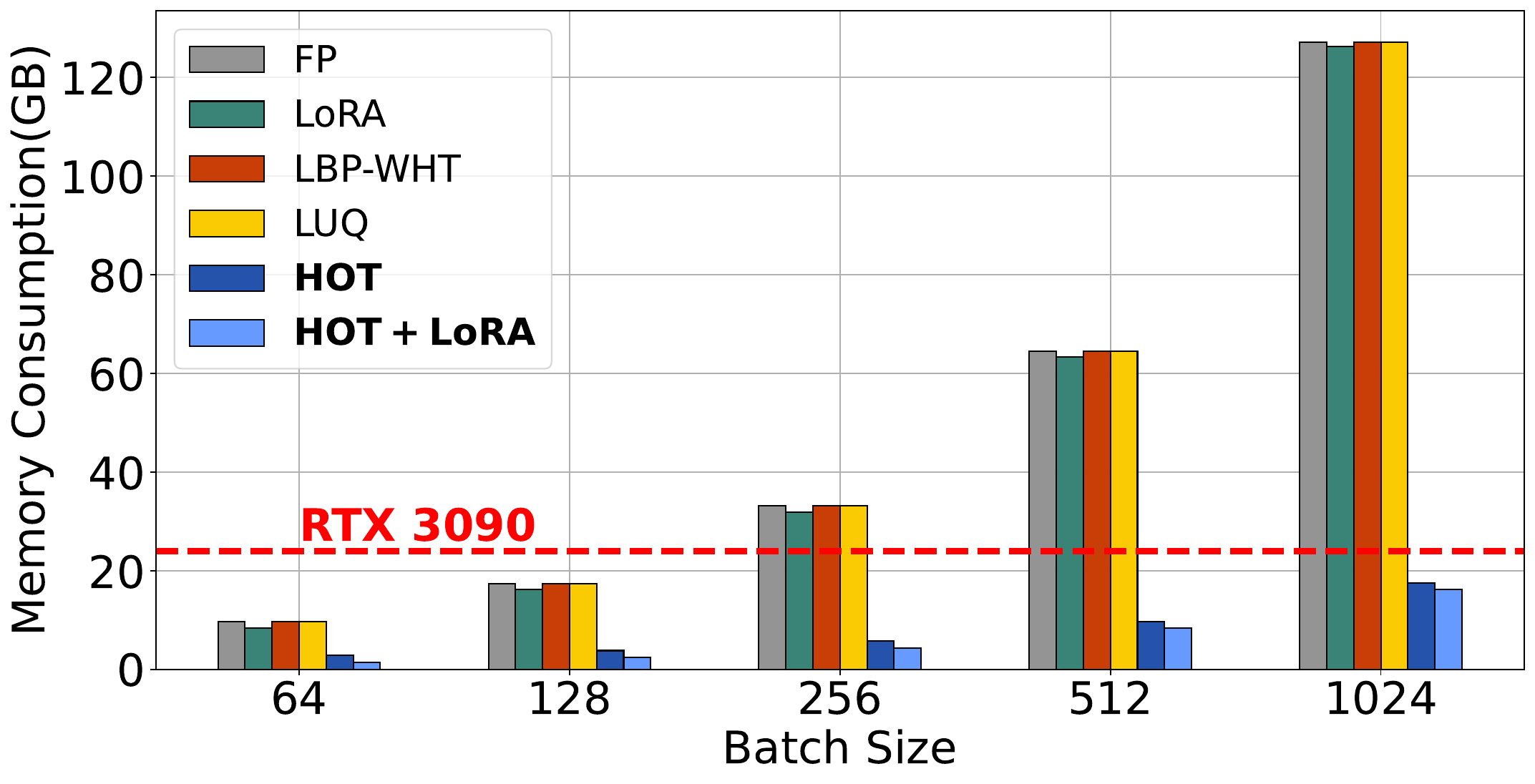}
    \caption{Memory requirements for training ViT-B~\cite{vit} on ImageNet-1k dataset~\cite{imagenet} with varying batch sizes. While FP and other efficienct BP methods fail to train with batch sizes of 256 and above on a single GPU having 24 GB memory, HOT enables training with batch sizes up to 1024.}
    \label{fig:max-batch-size}
\end{figure}

\begin{table}[]
\centering
\resizebox{\columnwidth}{!}{%
\begin{tabular}{ccccc}
\toprule
                                & \multicolumn{4}{c}{\textbf{Objectives}}                                                                      \\ \cline{2-5} 
                                & \makecell{Weight \& \\ Optimizer Memory} & \makecell{Intermediate\\Activation} & \makecell{Backpropagation\\Speed} & \makecell{Trained Model\\Quality} \\ \midrule 
\multicolumn{1}{c}{\textbf{LoRA}~\cite{lora}}       &         \textcolor[RGB]{0,150,0}{\checkmark}              &       \textcolor{red}{\ding{55}}                  &     \textcolor[RGB]{230,150,0}{\ding{115}}                   &           \textcolor[RGB]{0,150,0}{\checkmark}          \\
\multicolumn{1}{c}{\textbf{LBP-WHT}~\cite{lbp-wht}}        &          \textcolor{red}{\ding{55}}                 &        \textcolor{red}{\ding{55}}                 &       \textcolor[RGB]{0,150,0}{\checkmark}                &       \textcolor[RGB]{230,150,0}{\ding{115}}               \\
\multicolumn{1}{c}{\textbf{LUQ}~\cite{luq}}        &           \textcolor{red}{\ding{55}}                &       \textcolor{red}{\ding{55}}                  &    \textcolor[RGB]{230,150,0}{\ding{115}}                  &      \textcolor[RGB]{0,150,0}{\checkmark}                 \\
\multicolumn{1}{c}{\textbf{HOT}}        &          \textcolor{red}{\ding{55}}                 &      \textcolor[RGB]{0,150,0}{\checkmark}                   &        \textcolor[RGB]{0,150,0}{\checkmark}               &          \textcolor[RGB]{0,150,0}{\checkmark}             \\
\multicolumn{1}{c}{\textbf{HOT + LoRA}} &          \textcolor[RGB]{0,150,0}{\checkmark}                    &      \textcolor[RGB]{0,150,0}{\checkmark}                   &      \textcolor[RGB]{0,150,0}{\checkmark}                 &   \textcolor[RGB]{0,150,0}{\checkmark}  \\ \bottomrule  \bottomrule                  
\end{tabular}%
}
\caption{Comparison with previous works. HOT is the only method which achieves activation memory reduction and actual speedup altogether without compromising model quality.}
\label{tab:comp-prev}
\end{table}

In this work, we propose a novel method called Hadamard-based Optimized Training (HOT), which focuses on optimizing matrix multiplication—a major contributor to training overhead. First, We thoroughly analyze the backward characteristics of activations and weights. Based on this insight, we determine the optimal strategy for selectively applying Hadamard-based optimization techniques tailored to each unique characteristic. Combined with our two new ideas—Activation Buffer Compression (ABC) for activation memory reduction and Layer-wise quantizer Selection (LQS) for fine-quality quantization—the benefits of HOT are maximized. Additionally, we explore and optimize the integration of HOT with LoRA, creating a comprehensive solution for optimized training.

Our experiments validate HOT's efficiency and training quality on both vision and language models using large-scale datasets. As shown in \cref{fig:max-batch-size}, HOT achieves the best efficiency with large batch input. Our results indicate that even on an actual GPU, HOT achieves a \textbf{2.6$\times$ speedup} with custom CUDA kernels and a \textbf{75\% memory reduction}, all without \textbf{compromising training accuracy} than FP. 

\begin{figure}
    \centering
    \includegraphics[width=1.0\linewidth]{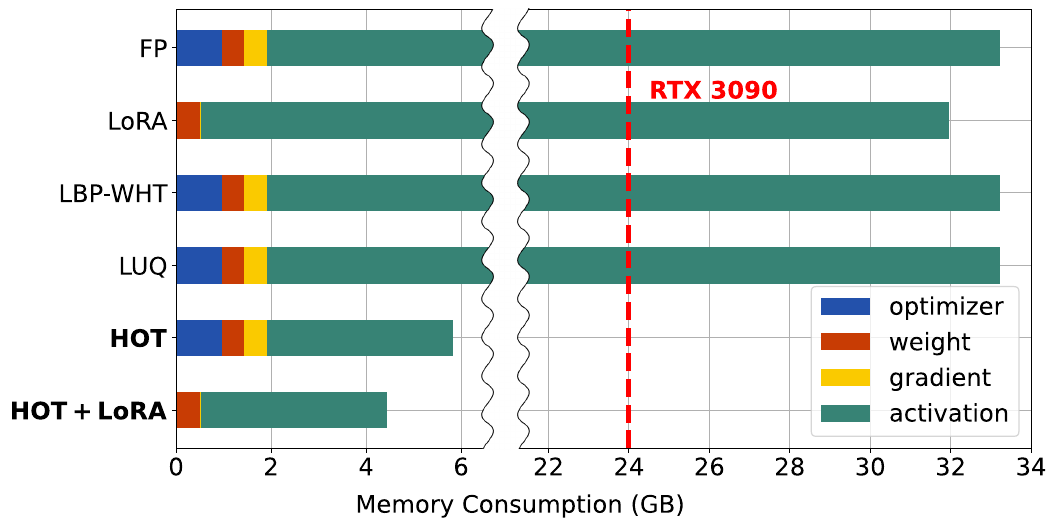}
    \caption{Component-wise memory consumption breakdown for different methods when training ViT-B~\cite{vit} on ImageNet-1k dataset~\cite{imagenet} with a batch size of 256.}
    \label{fig:figure1-breakdown}
\end{figure}

\section{Related Work}
\label{sec:related_work}
\subsection{Quantization for Efficient Training}
Numerous studies have explored the potential of quantization for efficient training. A common approach involves applying quantization to the forward propagation. Chmiel et al.~\cite{luq} proposed a logarithmic quantizer with a custom FP4 format for gradient optimization, though it had limitations in hardware acceleration. Xi et al.~\cite{xi2023training} combined forward quantization with structural pruning for gradients, demonstrating success in language tasks. Other approaches, such as FP8 training~\cite{micikevicius2022fp8, peng2023fp8lm} and integer-only training~\cite{wang2022niti, ghaffari2022integer}, aim to modify representations across the entire training process. \textbf{In contrast, our method maintains full precision in the forward propagation to ensure accurate loss evaluation and preserve training quality, while only quantizing the gradient computation for optimization.}

\subsection{Rank Reduction for Efficient Training}
Several approaches focus on reducing training costs through low-rank decomposition. LoRA~\cite{lora} reduces the fine-tuning cost of large language models (LLMs) by specifically targeting the memory usage of their massive weights. LBP-WHT~\cite{lbp-wht} was the first to introduce Hadamard low-rank approximation in backpropagation, achieving practical acceleration. However, it suffers from significant accuracy degradation on large-scale models. HOT also employs HLA, but we apply it selectively, taking into account the characteristics of backpropagation paths to minimize its drawbacks.
\section{Preliminaries}
\label{sec:preliminaries}
HOT builds upon two advanced optimization techniques—Hadamard Quantization (HQ) and Hadamard Low-rank Approximation (HLA)—and applies them selectively, taking into account the characteristics of the backpropagation for weight and activation gradient, respectively.

In this section, we introduce how HQ and HLA are leveraged for BP paths. Before we begin, we first define the notation for matrix multiplication. The operator $\cdot$, as used in this paper, denotes matrix-matrix multiplication between two matrices. Given two 2-dimensional matrices \( P \in \mathbb{R}^{M \times N} \) and \( S \in \mathbb{R}^{N \times K} \), their product \( R = P \cdot S \in \mathbb{R}^{M \times K} \) is obtained by taking inner products along the \( N \) dimension.

\begin{figure}
    \centering
    \small
    \includegraphics[width=1.0\linewidth]{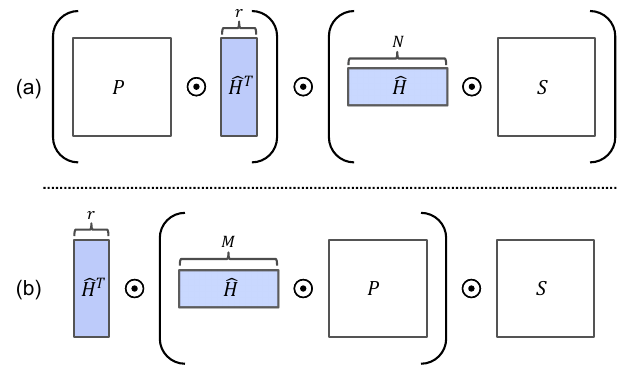}
    \caption{Illustration of (a) internal HLA and (b) external HLA. Internal HLA reduces dimension $N$ to $r$, while external HLA compress $M$ to $r$. The operator $\odot$ represents matrix multiplication.}
    \label{fig:HLA}
\end{figure}

\subsection{Hadamard Transformation}
\label{sec:Hadamard_Transformation}
The Hadamard Transformation (HT) \cite{sylvester1867lx} is the core building block for mitigating quality loss during quantization and low-rank approximation. HT can be viewed as a type of Fourier transformation, converting an \( n \)-dimensional vector \( v \in \mathbb{R}^{2^d} \) into the spectral domain \( \tilde{v} \in \mathbb{R}^{2^d} \) through the following linear transformation:
\begin{equation}
    \tilde{v} = H_d \cdot v,
\end{equation}
where $H_d$ is the Walsh-hadamard matrix, defined by following recurrence equation:
\begin{equation}
    \mathrm{H}_1 = \frac{1}{\sqrt{2}} \begin{bmatrix}
 1 & 1 \\ 
 1 & -1 
\end{bmatrix}, \quad \mathrm{H}_{n} = \mathrm{H}_1 \otimes \mathrm{H}_{n-1},
\end{equation}
where \(\otimes\) denotes the Kronecker product.

A key advantage of the HT is its computational efficiency: the Fast Walsh-Hadamard Transform \cite{shanks1969computation} requires only \(\mathcal{O}(2^d \cdot \log 2^d)\) operations, relying solely on additions and subtractions. Additionally, we apply a block-diagonal transformation, also known as the order-\(n\) 2D HT \cite{xi2023training}, which divides dimensions into specific tiles (typically 16) and performs transformations for each tile independently.

\subsection{Hadamard Quantization}
Recently, the combination of the HT with quantization, known as Hadamard Quantization (HQ), has gained attention for its lower degradation in low-precision settings. HQ applies HT along the dimension where inner products are computed, leveraging the orthogonality of the Hadamard matrix to ensure consistent outputs in subsequent updates:
\begin{equation}\label{eq:HT}
    R = P \cdot S= P\cdot (H^T \cdot H) \cdot S = (P\cdot H^T) \cdot (H \cdot S),
\end{equation}
$H \in R^{N \times N}$ is a dimension-matched block-diagonal Walsh-Hadamard matrix. For the arbitray quantization  $Q(\cdot)$, HQ  applies it to P and S after they are mapped to the frequency domain. In this case, the approximated output $\tilde{R}$ is calculated as follows:
\begin{equation}
    \tilde{R} = Q(P\cdot H^T) \cdot Q(H \cdot S).
\end{equation}
HT converts irregular data with outliers into frequency distributions, making the data more resilient to quantization \cite{xi2023training, schiemer2023hadamard}. Although HQ introduces minor computational overhead, the added cost is justified by its benefits.

\subsection{Hadamard Low-rank Approximation}
\label{sec:HLA}
The Hadamard Low-rank Approximation (HLA) is a technique that reduces computational and memory costs, as illustrated in \cref{fig:HLA}. HLA leverages HT to convert data to the frequency domain, selecting the low-frequency components after transformation \cite{lbp-wht}.

HLA can be categorized into internal and external HLA, depending on the dimension to which the low-rank approximation is applied. Internal HLA applies the HT to the inner dimension, approximating the operation by selecting only \( r \ll N \) dimensions. When we denote this reduced Walsh-Hadamard matrix as \( \hat{H} \), the approximated output \( \hat{R} \) is calculated as follows:
\begin{equation}
    \hat{R} = (P \cdot \hat{H}^T) \cdot (\hat{H} \cdot S).
\end{equation}

In contrast, external HLA applies low-rank approximation to the \( M \) dimension of \( P \) or the \( K \) dimension of \( S \). For example, when applying HLA to the \( M \) dimension, the approximated output is calculated as follows:
\begin{equation}
    \hat{R} = \hat{H}^T \cdot (\hat{H} \cdot P) \cdot S.
\end{equation}

Both methods offer computational and memory advantages by reducing dimensionality. Although HT introduces minor overhead, it is a manageable cost given the benefits. 

In the existing LBP-WHT~\cite{lbp-wht} paper, external HLA is used for lightweight computation along the \( g_x \) path, while internal HLA is applied when calculating \( g_w \). However, this approach results in notable accuracy loss. Our observation indicates that HLA is unsuitable for the activation gradient path, suggesting a need to explore alternative methods.
\section{Optimization Sensitivity Analysis}
\label{sec:Analysis}
For optimal acceleration and memory savings with minimal accuracy loss, we conduct an in-depth analysis of the optimization sensitivity in the BP of matrix multiplication. Although the two operands in backpropagation, \( g_x \) and \( g_w \), serve similar functions, their tensor properties differ significantly. Consequently, each path may exhibit different sensitivities to optimization.

\subsection{Notataion for BP }
In our explanation, we use unified notation for each dimension. Input channels are denoted as (I), output channels as (O), and the sequence dimension as (L). Although activations include a batch dimension \(B\), we omit it here for brevity. Based on this notation, in the forward pass, for input activation \( x \in \mathbb{R}^{L \times I} \) and weight \( w \in \mathbb{R}^{O \times I} \), the output \( y \in \mathbb{R}^{L \times O} \) is calculated as follows:

\begin{equation}
    y = x \cdot w^T
\label{equ:fwd}
\end{equation}

In the backward pass, given the gradient of the output activation \( g_y \in \mathbb{R}^{L \times O} \), we can calculate the gradients of the input activation \( g_x \in \mathbb{R}^{L \times I} \) and the weight \( g_w \in \mathbb{R}^{O \times I} \) using the chain rule:
\begin{equation}
g_w = g_y^T \cdot x, \quad g_x = g_y \cdot w.
\label{equ:bwd}
\end{equation}

\begin{table}[!t]
\centering
\small
\captionsetup{font={small}}
\setlength{\tabcolsep}{0.5pt}
\renewcommand{\arraystretch}{1.1}
\begin{tabular*}{\columnwidth}{@{\extracolsep{\fill}}ccc@{}}
\toprule
\textbf{$\mathbf{g_x}$ path}   & \textbf{$\mathbf{g_w}$ path}   & \textbf{Accuracy} \\ \toprule
FP           & FP           & 76.46             \\ \midrule
FP           & HT + 4-bit Q & 72.43             \\
FP           & Internal-HLA & {\ul \textbf{76.29}}            \\ \midrule
4-bit Q      & FP           & 73.4               \\
HT + 4-bit Q & FP           & {\ul \textbf{76.16}}             \\
External-HLA & FP           & 72.01             \\
Internal-HLA & FP           & 51.10             \\ \bottomrule \bottomrule 
\end{tabular*}
\caption{Table showing the application of HT and HLA with varying precision when pretraining ResNet50 with CIFAR100. A stochastic quantization~\cite{luq} is used as the quantizer.}
\label{tab:HT_analysis}
\vspace{-1mm}
\end{table}

The computation of each gradient incurs the same cost as the forward pass operations, so both paths should be optimized judiciously using appropriate methods. Notably, when optimizing Vision Transformers (ViT)~\cite{vit} or fully convolution layers, the same optimization techniques can be applied by substituting \( L = W \times H \), where \( W \) and \( H \) represent the spatial width and height, respectively.

\subsection{Activation Backpropagation Optimization}
\label{sec:gx_analysis}
The \( g_x \) path is computed as a matrix multiplication of the output gradient (\( g_y \)) and the weight (\( w \)), as shown in \cref{equ:bwd}. Since \( g_y \) is stored locally and \( w \) is persistent, optimization for \( g_x \) should prioritize acceleration over memory reduction.
Although \( g_x \) computations are performed across all layers for each instance, there is an opportunity for robustness in the batch dimension, as the instance-wise gradient is averaged over this dimension.

To optimize the \( g_x \) path, we analyzed three approaches: HQ, internal HLA, and external HLA, finding that HQ is the most effective. The existing SOTA, LBP-WHT, reduces computational costs by applying external HLA to the \( L \) dimension of \( g_y \). We also evaluated internal HLA by reducing the rank of the common \( O \) channel. However, as shown in Table \ref{tab:HT_analysis}, both HLA methods resulted in large accuracy degradation in the \( g_x \) path.

\begin{figure}[!t]
    \centering
    \includegraphics[width=\linewidth]{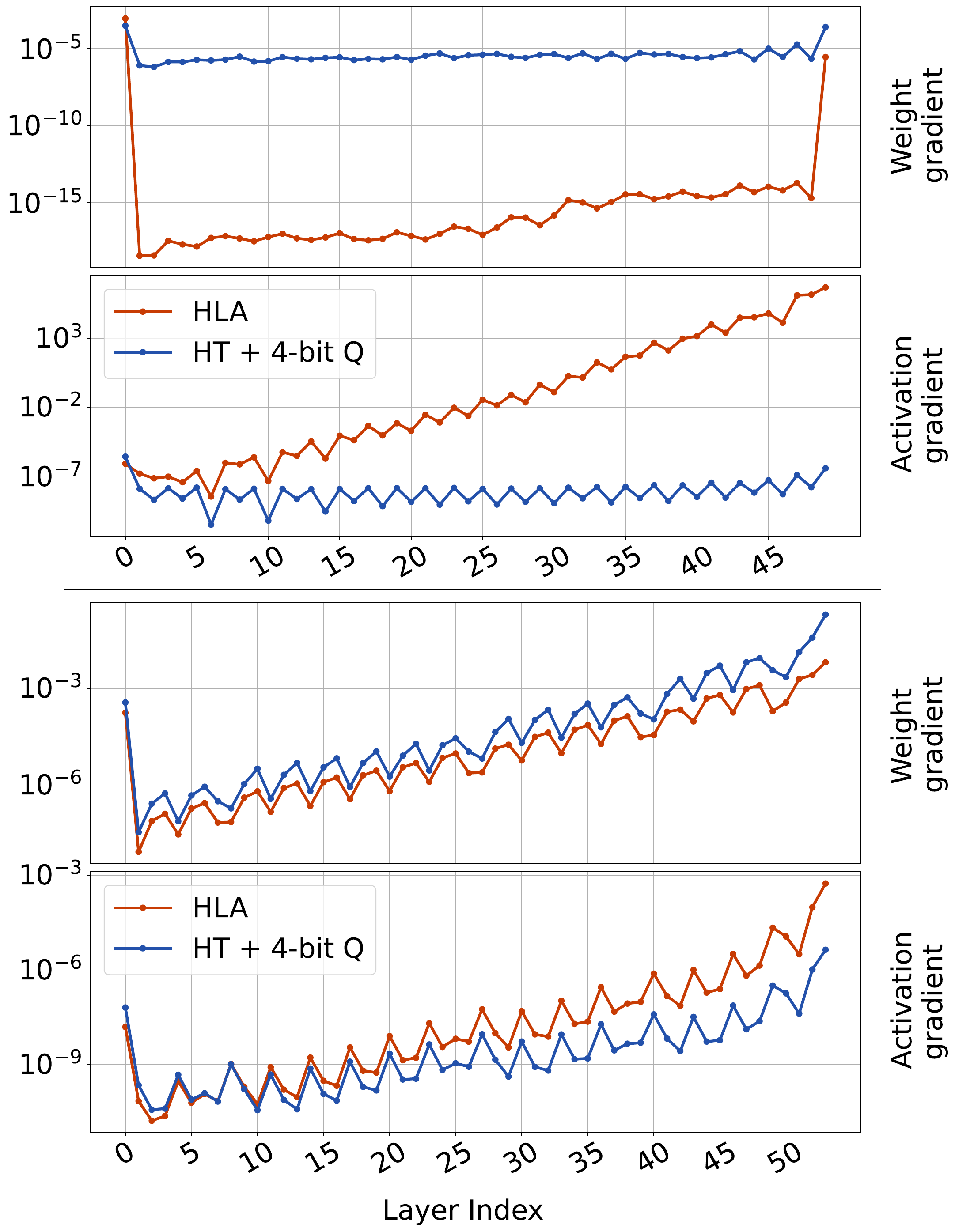} 
    \caption{Layer-wise MSE error analysis for ViT-B (top) and ResNet-50 (bottom). The graphs demonstrate higher errors in HT+INT4 for the weight gradient ($g_w$) path, while showing accumulated errors in HLA for the activation gradient ($g_x$) path.}
    \label{fig:Layer_per_Qerror}
    \vspace{-1mm}
\end{figure}

This damage seems to arise from frequency loss patterns generated during HLA. When these patterns accumulate across the network, they create trends difficult to correct, even with batch dimension averaging. Moreover, these errors worsen rapidly in deeper layers, as shown in \cref{fig:Layer_per_Qerror}.

Meanwhile, quantization can be modeled as adding random noise~\cite{baskin2021nice}, which is mitigated by averaging across the batch dimension. Moreover, applying HT plays a crucial role in error reduction. Notably, combining HT with 4-bit quantization performs surprisingly well and opens up opportunities for acceleration, as 4-bit GEMM can be over 3$\times$ faster than FP16 GEMM via Tensorcore acceleration~\cite{wu2023understanding}.

\subsection{Parameter Backpropagation Optimization}
\label{sec:gw_analysis}
In BP, \( g_w \) is computed as the product of intermediate activations and the transpose of \( g_y \)~\cref{equ:bwd}. Since \( g_w \) stores and reuses activations from the forward pass, reducing its memory consumption is a key optimization goal.

After analyzing the characteristics of \( g_w \), we confirmed that internal HLA is highly suitable, as shown in \cref{tab:HT_analysis}. Fundamentally, \( g_w \) updates are performed by taking the averaged value across the \( L \) dimension, which acts as a low-pass filter in the frequency domain~\cite{lbp-wht}. When performing internal HLA, the accuracy loss from selecting only low-frequency ranks for the \( L \) dimension is mitigated, as it aligns with this filtering effect.

In contrast, low-bit quantization caused significant accuracy degradation, even with HT applied. This is likely due to weight gradients directly influencing the weight update trajectory through repeated accumulation, making quantization errors immediately impactful on quality. Observing $g_w$  in \cref{fig:Layer_per_Qerror}, we can see that HQ induces larger errors than HLA across $g_w$ paths in all models. While quantization could be considered for $g_w$ to manage memory consumption, it must be applied conservatively.

Through additional observations, we identified a potential approach to further mitigate the quantization sensitivity of $g_w$. In Transformer-based architectures, as shown in \cref{fig:gradient_tensor_case}, we observed gradient outliers in certain layers. When gradients generated from specific tokens have unusually large values compared to others, applying per-tensor quantization results in significant errors. Instead, by applying per-token quantization~\cite{xiao2023smoothquant} based on token-specific statistics, we can greatly reduce quantization errors. This approach is practically meaningful: because per-token quantization maintains a consistent scale across the $L$ dimension, we can still utilize INT8 GEMM, with the scaled output achieved by multiplying the token-wise scale with the GEMM output. This new approach allows us to implement additional optimizations in the $g_w$ path, alongside HLA. 

\begin{figure*}[!t]
    \centering
    \includegraphics[width=\linewidth]{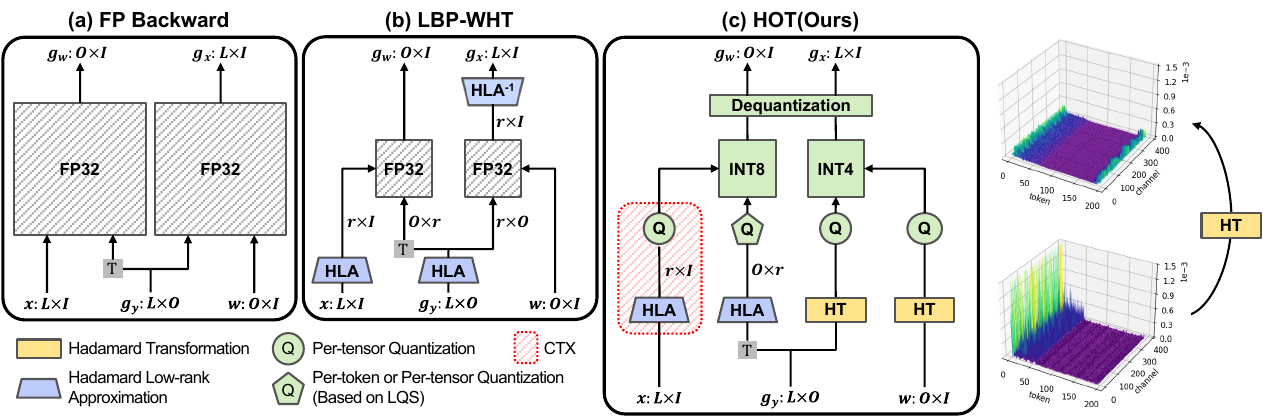} 
    \caption{The pipelines for (a) standard BP, (b) LBP-WHT~\cite{lbp-wht}, and (c) HOT. HOT reduces memory consumption by compressing activations for BP using HLA and INT8 quantization, storing them in CTX (shown in red in (c)). It accelerates computations through integer matrix multiplication for gradient calculations (represented by INT8 and INT4 rectangular sections in (c)).}
    \label{fig:main_figure}
    \vspace{-4mm}
\end{figure*}
\section{HOT: Hadamard-based Optimized Training}
Based on our previous observations, we propose a holistic optimization pipeline, HOT: Hadamard-based Optimized Training. HOT employs optimal strategies tailored to the characteristics of both the $g_w$ and $g_x$ paths. Furthermore, for the $g_w$ path, we introduce Activation Buffer Compression (ABC), which achieves memory savings, and Layer-wise Quantizer Selection (LQS), which balances stable quantization and training speed.

\subsection{Optimization on activation gradient path}
\label{sec:gx_path}
HOT applies only INT4 quantization to $g_x$ without low-rank approximation. First, HT is applied to $g_y$ along the $O$ dimension to generate the projected tensor, after which quantization is applied to the projected $g_y$. For $w$, HT is similarly applied along the $O$ dimension to produce the projected $w$, which is then quantized to INT4. Since both $g_y$ and $w$ are quantized to integers, the integer $g_x$ can be efficiently computed through INT4 GEMM and dequantized to FP32.

We accelerate computation through INT4 matrix multiplication using NVIDIA TensorCores. Our optimization combines a custom CUDA implementation and operator fusion for HT and quantization to maximize acceleration.

For quantization, unbiased quantization is essential; biased quantization can significantly degrade training quality \cite{chmiel2022accurate}. It is known that stochastic rounding provides an unbiased estimate \cite{chen2020statistical}, so HOT employs min-max stochastic quantization. However, to minimize the overhead of generating random numbers, we use a pseudo-stochastic quantizer \cite{wang2022niti}, leveraging the lower 11 bits of FP32 as pseudo-random values to determine whether to round FP numbers.

\subsection{Optimization on weight gradient path}
For $g_w$, HOT combines HLA and INT8 quantization. As mentioned in section \ref{sec:gw_analysis}, this reflects the characteristic that the $g_w$ computation path is robust to HLA but sensitive to quantization.

The computation process is as follows. First, $g_y$ is transposed and HLA is applied along the L dimension. Next, INT8 quantization is applied to the compressed $g_y$. For $x$, this operation is also applied along the L dimension to generate compressed $x$, which is then quantized to 8-bit integers. Since both $g_y$ and $x$ are quantized to 8 bits, integer $g_w$ is calculated through INT8 matrix multiplication. Finally, integer $g_w$ is dequantized to FP32.

\subsubsection{Activation Buffer Compression} 
Additionally, reducing activation memory is a critical factor in the $g_w$ path. To address this, we introduce Activation Buffer Compression (ABC) to minimize memory costs.

To further improve memory efficiency in the $g_w$ path, HLA and quantization for $x$ can be applied immediately after the forward pass, rather than during the backward pass. After the forward pass, when storing $x$ in main memory for the backward pass of each layer, we can significantly reduce activation memory usage by storing $x$ with HLA and 8-bit quantization already applied. This technique allows activation size to be compressed by up to 50\% during the HLA process, with an additional compression of 25\% achieved during quantization when converting from FP32 to INT8. This theoretically enables memory savings of up to 12.5\% compared to the original. We apply it as a fundamental memory optimization strategy.

\subsubsection{Layer-wise Quantizer Selection}
To further reduce quantization errors in $g_w$, per-token quantization can be applied to $g_y$. However, this approach involves the trade-off of additional inference costs due to the need for token-wise scale calculation. Since the characteristics of $g_y$ vary by layer, HOT can selectively apply either per-token or per-tensor quantization.

\begin{figure}[!t]
    \centering
    \small
    \includegraphics[width=\linewidth]{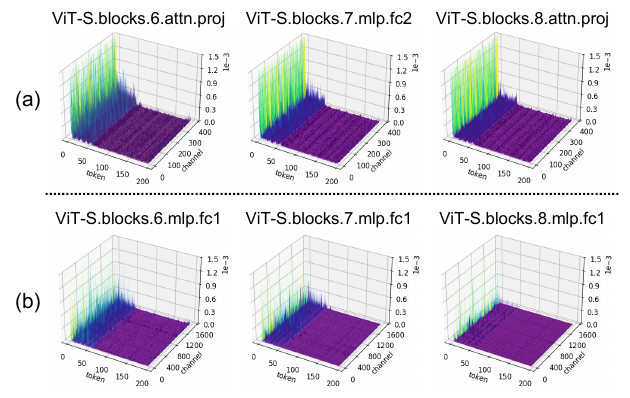} 
    \caption{The illustration for output gradient tensor of (a) attention proj layer of the 6, 8th block and fc2 layer of the 7th block in ViT-S~\cite{vit}; (b) fc1 layers of the 6th, 7th and 8th blocks in ViT-S. While (a) benefits from per-token quantization, (b) shows comparable performance with per-tensor quantization.}
    \label{fig:gradient_tensor_case}
\end{figure}

By analyzing the patterns of \( g_y \) across layers, we identified two distinct cases: (a) layers that consistently show gradient outliers for specific tokens, and (b) layers where gradient outliers are difficult to recognize. For example, as shown in Figure \ref{fig:gradient_tensor_case}, the first case consistently appears in the attention projection and fc2 layers of ViT-S~\cite{vit}, while the second case occurs consistently in its fc1 layer. In Case (a), tensor-wise quantization results in significant accuracy loss, so token-wise quantization should be applied despite its associated overhead. In Case (b), tensor-wise quantization is preferable, as the computational overhead of token-wise quantization outweighs its accuracy benefits.

To implement this, HOT uses a hand-crafted, empirical selection process by performing a backward pass on a small calibration set prior to training. Specifically, for each layer’s $g_y$, we calculate the Mean Squared Error (MSE) between FP and per-token quantization, and between FP and per-tensor quantization. If the error difference is less than 50\%, per-tensor quantization is applied; if the difference is 50\% or greater, per-token quantization is used. We refer to this layer-wise quantizer selection method as LQS. Although the trade-offs require further detailed evaluation, this straightforward strategy provides satisfactory benefits in both computational throughput and training quality.

\subsection{Joint optimization with LoRA}
LoRA~\cite{lora} and HOT have distinct optimization objectives, making them complementary in application. We have confirmed that by applying LoRA in the forward pass and HOT in the backward pass, we can effectively compress model weights and optimizer states through LoRA, while compressing intermediate activations through HOT.

The key characteristic of using HOT and LoRA together is the differentiated approach to its frozen weight and decomposed weight. In frozen weight, where parameters remain unchanged, HOT skips the calculation of $g_w$ and only computes $g_x$ for the backward pass to the next layer. 

In contrast, within LoRA's decomposed weight, updates are applied to the decomposed weights after the backward pass. We found that performing traditional BP without applying HOT in this part is crucial for preserving accuracy. Since LoRA already reduces the cost of $g_w$, the additional cost reduction from HOT in these regions is minimal, making this trade-off beneficial. By optimally combining the strengths of LoRA and HOT, we have developed a holistic and efficient training method that maximizes training performance while minimizing memory usage.

\begin{table*}[!t]
    \centering
    \setlength{\tabcolsep}{10pt}
    \renewcommand{\arraystretch}{1.2}
    \captionsetup{font={small}}
    \vspace{-0.1cm}

    \subfloat[\textbf{Fine-tuning on classification task}]{
        \scalebox{0.85}{
            \begin{tabular}{cc|cccccc@{}}
                \toprule
                \textbf{Dataset} & \textbf{Model} & \textbf{FP} & \textbf{LoRA}~\cite{lora} & \textbf{LUQ}~\cite{luq} & \textbf{LBP-WHT}~\cite{lbp-wht} & \textbf{HOT} & \textbf{HOT + LoRA} \\ \midrule
                \multirow{6}*{CIFAR10~\cite{cifar}}    & EfficientNetV2-s~\cite{effnetv2}       & 97.61 & -     & NaN & 94.09 & \textbf{\uline{96.79}} & - \\
                           & EfficientNetV2-m~\cite{effnetv2}       & 98.01 & -     & NaN & 95.85 & \textbf{\uline{96.17}} & - \\
                           & EfficientFormer-L1~\cite{effformer}     & 97.31 & 94.14 & 96.42 & 94.6  & \textbf{\uline{96.49}} & 94.01 \\
                           & EfficientFormer-L7~\cite{effformer}     & 98.65 & 97.10 & 97.86 & 97.22 & \textbf{\uline{98.03}} & 96.90 \\
                           & ViT-S~\cite{vit}                  & 98.68 & 98.70 & 97.37 & 97.24 & 98.21 & \textbf{\uline{98.60}} \\
                           & ViT-B~\cite{vit}                  & 96.36 & 96.07 & 95.70 & 91.37 & 95.01 & \textbf{\uline{96.03}} \\[1mm]
                           \midrule
                \multirow{6}*{CIFAR100~\cite{cifar}}   & EfficientNetV2-s ~\cite{effnetv2}      & 87.36 & -     & NaN & \textbf{87.10} & 86.29 & - \\
                           & EfficientNetV2-m~\cite{effnetv2}       & 84.89 & -     & NaN & 55.08 & \textbf{\uline{83.91}} & - \\
                           & EfficientFormer-L1~\cite{effformer}     & 85.73 & 84.84 & 78.70 & 83.27 & \textbf{\uline{84.07}} & 83.42 \\
                           & EfficientFormer-L7~\cite{effformer}     & 88.04 & 85.09 & 76.33 & 85.10 & \textbf{\uline{87.39}} & 85.01 \\
                           & ViT-S~\cite{vit}                  & 86.81 & 85.67 & \textbf{86.37} & 85.96 & 85.81 & 85.49 \\
                           & ViT-B~\cite{vit}                  & 93.45 & 92.61 & 91.76 & 92.47 & \textbf{\uline{92.99}} & 92.51 \\
                \bottomrule \bottomrule 
            \end{tabular}
        }
    }
    \vspace{0.1cm}

    \subfloat[\textbf{Fine-tuning on semantic segmentation and object detection}]{
        \scalebox{0.85}{
            \begin{tabular}{@{}cc|cccccc@{}}
                \toprule
                \textbf{Dataset} & \textbf{Model} & \textbf{FP} & \textbf{LoRA}~\cite{lora} & \textbf{LUQ}~\cite{luq} & \textbf{LBP-WHT}~\cite{lbp-wht} & \textbf{HOT} & \textbf{HOT + LoRA} \\ \midrule 
                VOC 2012~\cite{voc}   & \multirow{2}*{Segformer-mit-b2~\cite{segformer}}      & 80.93 & 79.40 & 78.70 & 78.40 & 78.86 & \textbf{\uline{79.10}} \\
                Cityscape~\cite{cityscape}  &        & 72.16 & 70.39 & 70.31 & 71.16 & \textbf{\uline{71.72}} & 70.37 \\[1mm]
                \midrule
                VOC 2007~\cite{voc}   & YOLO V5-S~\cite{yolov5}              & 85.80 & -     & NaN   & 85.00 & \textbf{\uline{85.10}} & - \\
                \bottomrule \bottomrule 
            \end{tabular}
        }
    }
    \caption{Quality degradation comparison of BP optimization techniques across classification and semantic segmentation tasks. HOT demonstrates superior performance over other methods across almost all models and maintains comparable accuracy when integrated with LoRA~\cite{lora}.}
    \label{tab:imagenet}
    \vspace{-4mm}
\end{table*}

\begin{table}[t]
\centering
\begin{adjustbox}{{width=\columnwidth}}
\begin{tabular}{cc|cccccc}
\toprule
\textbf{Dataset} & \textbf{Model} & \textbf{FP} & \textbf{LoRA}~\cite{lora} & \textbf{LUQ}~\cite{luq} & \textbf{LBP-WHT}~\cite{lbp-wht} & \textbf{HOT} & \textbf{\makecell{HOT + \\ LoRA}} \\ 
\midrule
  MRPC~\cite{glue} & BERT-base~\cite{bert}   & 86.52 & 81.61 & \textbf{84.56} & 80.88 & 84.31 & 81.6 \\
  Alpaca~\cite{alpaca} & Llama3-8B~\cite{llama3}  & 3.28 & 3.8 & NaN & NaN & \textbf{\uline{3.29}} & 3.78 \\ 
\bottomrule \bottomrule 
\end{tabular}
\end{adjustbox}
\caption{Performance comparison of MRPC accuracy(\%) and Alpaca perplexity(\%) in LLM fine-tuning. Higher MRPC accuracy and lower Alpaca perplexity indicate better performance.}
\label{tab:LLM_experiment}
\vspace{-4mm}
\end{table}

\section{Experiment}
\label{sec:experiment}

We conducted comprehensive experiments to validate the superiority of the proposed technique. We set FP, LUQ~\cite{luq}, LBP-WHT~\cite{lbp-wht}, and LoRA~\cite{lora} as comparison groups, with LoRA results reported only for models with attention layers. Unless otherwise specified, ABC and LQS were applied by default. Detailed experimental settings including low-pass vector selection criteria of HLA and hyper-parameters are described in Supplementary Materials.

\subsection{Efficient fine-tuning on vision tasks}

First, we compared fine-tuning quality across representative vision tasks: Classification, object detection, and semantic segmentation, with results shown in table~\ref{tab:imagenet}. For classification, we conducted experiments on EfficientNetV2~\cite{effnetv2}, EfficientFormer~\cite{effformer}, and ViT~\cite{vit}, fine-tuning models pretrained on ImageNet-1k~\cite{imagenet} using CIFAR10 and CIFAR100 datasets~\cite{cifar}. For object detection and semantic segmentation, we conducted experiments using Segformer~\cite{segformer} and YOLO-V5~\cite{yolov5} models respectively, fine-tuning models pretrained on COCO dataset~\cite{cocodataset} using PASCAL VOC~\cite{voc} and Cityscape~\cite{cityscape} datasets.

As shown in table~\ref{tab:imagenet}, HOT demonstrates performance significantly better than previous SOTA in fine-tuning. Notably, LUQ, which is a former state-of-the-art method, fails to train on certain architectures, but our method achieves stable fine-tuning regardless of the architecture. It also shows superior quality in almost all cases compared to LBP-WHT. Additionally, we confirmed that combining with LoRA proceeds successfully with minimal degradation.

\subsection{Fine-tuning result of Large Language Model}

We extended the scope beyond vision models to LLM fine-tuning to validate its effectiveness. Table \ref{tab:LLM_experiment} shows the results of fine-tuning BERT-base~\cite{bert} and Llama3-8B~\cite{llama3}, two representative language models, on MRPC~\cite{glue} and Alpaca~\cite{alpaca} datasets respectively.

The experimental results with Llama3-8B particularly demonstrate superiority of HOT. State-of-the-art BP optimization techniques like LUQ and LBP-WHT showed severe degradation as model depth increased, sometimes resulting in failure. In contrast, HOT achieved stable training even with the deep architecture of Llama3-8B. In fine-tuning the relatively smaller BERT-base, HOT demonstrated performance comparable to existing SOTA methods.

\subsection{Performance Analysis}
A key factor of HOT is actual acceleration while reducing memory footprint. In this section, we compare the performance advantages of HOT against existing methods.

\subsubsection{Memory Consumption}

HOT dramatically reduces memory usage through activation compression using ABC. In its basic implementation, HOT shows memory efficiency similar to LBP-WHT~\cite{lbp-wht}. However, with the introduction of ABC, it successfully achieves an additional 75\% memory reduction, further emphasizing its effectiveness in managing memory. 

\begin{table}[t] 
   \centering
   
   \begin{adjustbox}{{width=\columnwidth}}
    \begin{tabular}{cc|cccc}
    \toprule
    \textbf{Dataset}                      & {\textbf{Model}} & \textbf{FP} & \textbf{LUQ}~\cite{luq} & \textbf{LBP-WHT}~\cite{lbp-wht} & \textbf{HOT}  \\ \midrule                                     
    \multirow{2}{*}{ViT-B~\cite{vit}}              & ImageNet-100~\cite{imagenet}                             & 77.87         & 75.97         &  54.25              & \textbf{\uline{77.31}}                \\
                                        & ImageNet-1k~\cite{imagenet}                             & 70.01          & 67.1        & NaN           & \textbf{\uline{69.4}}                    \\ 
    \bottomrule \bottomrule
    \end{tabular}
    \end{adjustbox}
    \caption{Accuracy results for pre-training tasks of ViT-B model on ImageNet-1k and ImageNet-100 datasets. Unlike other methodologies, HOT shows less than 1\% degradation with FP.}
    \label{tab:pretraining}
    \vspace{-5mm}
\end{table}

Figure \ref{fig:mem_comp} shows memory usage analysis for ResNet-50~\cite{resnet}, ViT-B~\cite{vit}, and EfficientFormer-L7~\cite{effformer} models with a batch size of 256 in ImageNet~\cite{imagenet}. Notably, while LBP-WHT and LUQ~\cite{luq} consume the same memory as FP32, HOT achieves up to 86\% memory reduction in ResNet-50 and up to 75\% in ViT models. Memory optimization of HOT makes it possible to train with a single NVIDIA RTX 3090 having 24GB main memory, which has  practical point in accessibility of deep learning training.

\subsubsection{Computation Cost}
 
While HOT incurs slight overhead compared to basic BP in transformation and quantization, it significantly reduces overall computation cost through low-precision operations. The overhead analysis is described in Supplementary Materials. Figure \ref{fig:mem_comp} shows the comparison of bit operations (bops)~\cite{baskin2021uniq, shin2023nipq} across ResNet-50~\cite{resnet}, ViT-B~\cite{vit}, and EfficientFormer-L7~\cite{effformer}. Specifically, HOT achieves approximately 64\% reduction in computational cost compared to FP in ResNet-50, which is more efficient than both LBP-WHT~\cite{lbp-wht} and LUQ~\cite{luq}. Similarly, it recorded 65\% reduction in both EfficientFormer-L7 and ViT-B.

We also measured the actual latency of BP through CUDA kernels on an RTX 3090 GPU, which is shown in Table \ref{tab:GPU_latency}. Experimental results show that HOT achieves up to 3.3$\times$ speedup in the fc2 layer of ViT-B, 2.6$\times$ in average of all layers of the model. The implementation detail of CUDA kernels is detailed in Supplementary Materials.

\begin{figure}[t]
     \centering
         \includegraphics[width=1.0\columnwidth]{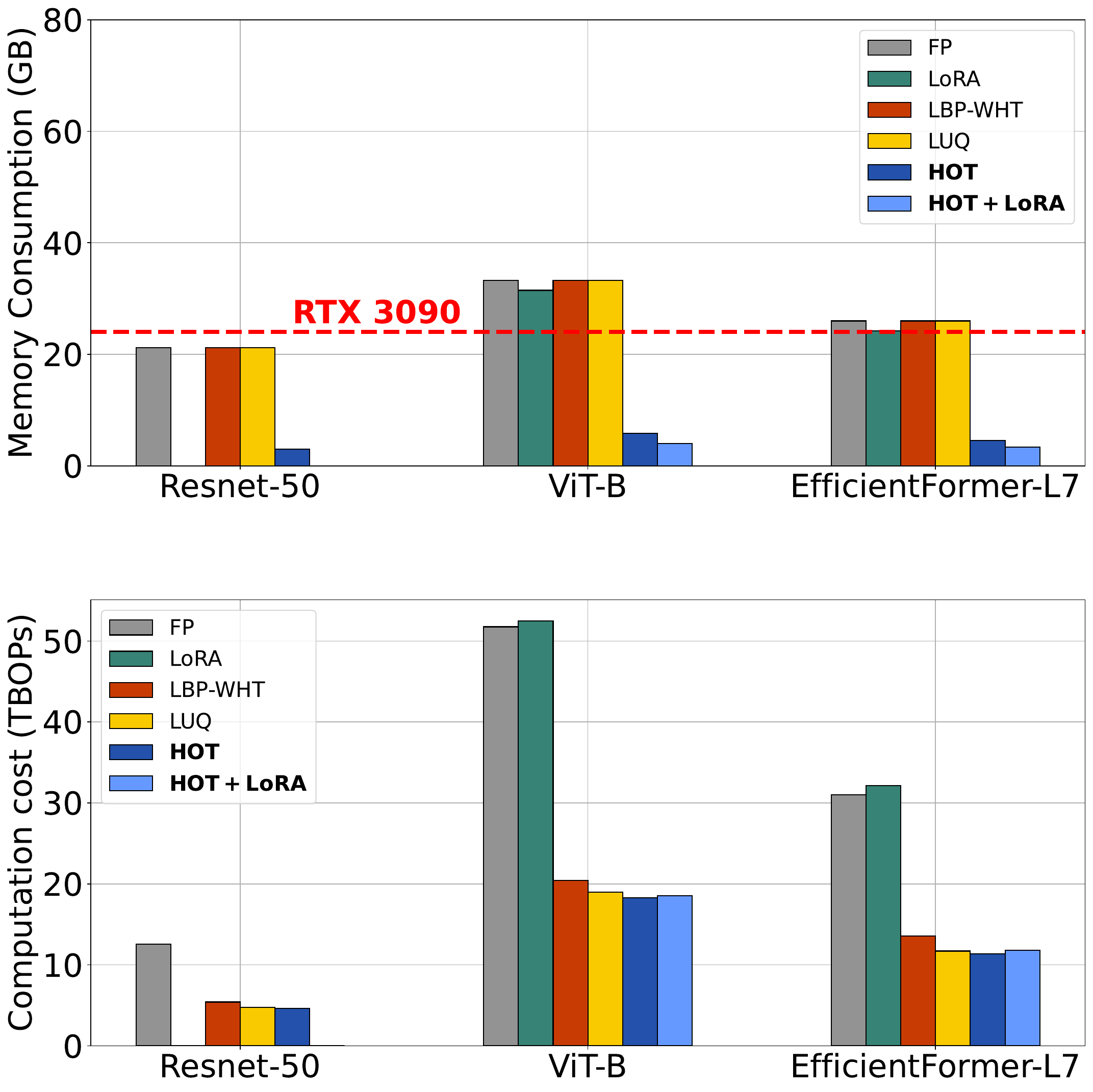}
     \caption{Estimated memory usage with 256 batch of ImageNet dataset~\cite{imagenet} and computational costs when applying HOT and existing methods across various models. HOT effectively reduces activation memory footprint, making training feasible on a single RTX 3090 GPU by significantly lowering resource requirements.}
     \label{fig:mem_comp}
     \vspace{-4mm}
\end{figure}

\subsection{Pre-training Result}
We evaluated the performance in the more challenging task of full training. Full training is known to be more vulnerable to gradient quantization or low-rank approximation as it starts from randomly initialized weights. 

Table \ref{tab:pretraining} shows the results of full training ViT-B~\cite{vit} model on ImageNet-1k and ImageNet-100 datasets~\cite{imagenet}. In ImageNet training, while other efficiency methods either show substantial degradation (LUQ) or train failure (LBP), HOT achieves performance nearly equivalent to FP. The experimental results show that HOT achieved superior performance compared to existing methods even under these demanding conditions, demonstrating stable convergence throughout the training. More detailed experimental results and analysis can be found in Supplementary Materials.

\begin{table}[!t] 
    \setlength{\tabcolsep}{0.3em}
    \small
    \captionsetup{font={small}}
    \centering
    \setlength{\tabcolsep}{1pt}
    \vspace{-1mm}
    \begin{tabular*}{\columnwidth}{@{\extracolsep{\fill}}lcccc@{}}
    \toprule
    \multicolumn{1}{c}{\textbf{(L, O,   I)}} & \textbf{Name} & \textbf{FP} & \textbf{LBP-WHT}& \textbf{HOT} \\ \midrule
    \multicolumn{5}{c}{\textbf{ResNet-50}}                                                        \\ \midrule
    (3136, 64,   256)    & layer1.conv1  & 115         & 106              & 62 (1.9$\times$)           \\
    (3136, 64,   576)    & layer1.conv2  & 134         & 117              & \textbf{\uline{59 (2.3$\times$)}}           \\
    (784, 128,   512)    & layer2.conv1  & 117         & 99               & 67 (1.8$\times$)          \\
    (784, 128,   1152)   & layer2.conv2  & 124         & 81               & 60 (2.1$\times$)          \\
    (196, 256,   2304)   & layer3.conv2  & 114         & 85               & 64 (1.8$\times$)          \\
    (49, 512,   4608)    & layer4.conv2  & 137         & 102              & 72 (1.9$\times$)          \\ \midrule
    \multicolumn{5}{c}{\textbf{ViT-B}}                                                            \\ \midrule
    (197, 2304,   768)   & qkv           & 182         & 110              & 70 (2.6$\times$)          \\
    (197, 768,   768)    & proj          & 122         & 108              & 71 (1.7$\times$)          \\
    (197, 3072,   768)   & fc1           & 226         & 120              & 73 (3.1$\times$)          \\
    (197, 768,   3072)   & fc2           & 233         & 112              & \textbf{\uline{72 (3.3$\times$)}}          \\ \midrule
    \multicolumn{5}{c}{\textbf{EfficientFormer-L7}}                                               \\ \midrule
    (3136, 384,   96)    & stages.0.fc1  & 125         & 123              & 63 (2.0$\times$)          \\
    (784, 768,   192)    & stages.1.fc1  & 129         & 108              & 68 (1.9$\times$)          \\
    (196, 1536,   384)   & stages.2.fc1  & 126         & 102              & 66 (1.9$\times$)          \\
    (49, 1536,   768)    & stages.3.qkv  & 128         & 105              & 62 (2.1$\times$)          \\
    (49, 768,   1024)    & stages.3.proj & 111         & 105              & 69 (1.6$\times$)          \\
    (49, 3072,   768)    & stages.3.fc1  & 146         & 110              & \textbf{\uline{66 (2.2$\times$)}}          \\ \bottomrule
    \end{tabular*}
    \caption{Latency($\mu$s) profiling results measured on RTX 3090 GPU with varying layer dimensions. HOT achieves significant speedups across different layers and architectures, outperforming LBP-WHT~\cite{lbp-wht} by large margin.}
    \label{tab:GPU_latency}
    \vspace{-1mm}
\end{table}

\label{sec:conclusion}
\section{Limitation and Broader impact}
As model size and depth increase, HOT tends to show more performance degradation. It is a clear limitation of HOT, despite incurring fewer errors compared to existing methods. Nevertheless, HOT can still be considered an optimal solution for training in resource-constrained environments with limited memory and computational capacity.

\section{Conclusion}
In this paper, we introduce HOT, a novel training methodology that reduces memory usage, speeds up computation, and maintains quality. We accelerate the activation gradient path using low-precision HQ while enhancing training stability in the weight gradient path through high-precision HLA quantization. We also improve memory efficiency by compressing activations with ABC and stabilize training with layer-wise quantizers (LQS). As a result, HOT achieves state-of-the-art performance across various vision and language tasks, delivering both high memory efficiency and significant acceleration.

\section*{Acknowledgement} 
This work was supported by IITP grant funded by the Korea government (MSIT) (Nos. RS-2019-II191906, RS-2023-00228970, RS-2024-00396013, RS-2024-00457882).
\clearpage
\setcounter{page}{1}
\maketitlesupplementary

\appendix 

\section{Overview}
This supplementary material provides detailed configurations of our experiments, ablation studies, visualization results, and pre-training experiments across various tasks and datasets.

We provide the following items:

\begin{itemize}
    \item{Detailed hyper-parameters used in our experiments are presented in \cref{sec:Appendix_Detailed_Experimental}.}

    \item{Incremental ablation studies on ABC and LQS, the subordinate methods of HOT, are discussed in \cref{sec:Appendix_incremental_ablation}.}

    \item{Ablation study on the number of low-pass ranks in HLA is presented in \cref{sec:Appendix_ablation_rank_selection}.}
    
    \item{Ablation study on the combination of HOT and LoRA is detailed in \cref{sec:Appendix_LoRA_ablation}.}
    
    \item{Theoretical analysis of the computational overhead introduced by HOT is provided in \cref{sec:Appendix_overhead_calculation}.}
    
    \item{Experimental results of pre-training across various tasks and datasets are presented in \cref{sec:Appendix_pretraining}.}
    
    \item{Implementation details of CUDA kernels and latency breakdown are documented in \cref{sec:Appendix_CUDA_kernel}.}
    
    \item{Additional visualization results of gradient tensor outlier patterns are presented in \cref{sec:Appendix_gradient_tensor_case}.}
\end{itemize}

\section{Detailed Experimental Settings}
\label{sec:Appendix_Detailed_Experimental}
In this section, we explain the detailed Hadamard Low-rank Approximation (HLA) configuration of low-pass vector selection and hyper-parameter setting for experiment of each task. All experiments except pre-training are conducted for fine-tuning task. 

\textbf{Hyper-paramemter of HLA:} We select top 8 low-pass vectors ($r = 8$) based on $LP_{L1}$ criteria of LBP-WHT\cite{lbp-wht} for our experiment. The ablation study of rank selection is explained on \cref{sec:Appendix_ablation_rank_selection}. The $LP_{L1}$ is a method for selecting low-pass vectors that simultaneously reflect the frequencies of both vertical and horizontal components of the image. Unlike the conventional sequency order of Hadamard matrix basis that only reflects horizontal components, this approach can effectively filter low-frequency components of image patches from both directions.

\textbf{Hyper-parameter of experiments: }For the classification tasks, we employed three different models: EfficientNetV2\cite{effnetv2}, EfficientFormer\cite{effformer}, and Vision Transformer (ViT)\cite{vit}. EfficientNetV2 was trained with a batch size of 64 and learning rate of 0.001, while both EfficientFormer and ViT utilized a batch size of 128 and learning rate of 0.00025. All models were trained for 50 epochs using the AdamW\cite{adamw} optimizer and cosine annealing scheduler\cite{cosineschduler}, implemented via the Timm library\cite{rw2019timm}.

For object detection, we implemented two approaches. The Segformer-mit-b2\cite{segformer} model from HuggingFace Transformers\cite{huggingface} was trained with a learning rate of 6e-5 for 50 epochs. YOLO-V5\cite{yolov5}, implemented using the official repository, was trained for 40 epochs with a learning rate of 0.00334, using SGD optimizer and linear decay scheduling with warmup.

The language model fine-tuning experiments involved BERT-base and LLama3-8B\cite{llama3}. BERT-base was trained using HuggingFace Transformers with a batch size of 32, maximum sequence length of 128, and learning rate of 2e-5 for 5 epochs. LLama3-8B, used a smaller batch size of 2 with maximum sequence length of 1024, maintaining the same learning rate for 4 epochs. Both models employed AdamW optimizer and cosine annealing scheduler.

For pre-training, we experimented with ResNet\cite{resnet}, EfficientFormer, and ViT models. ResNet was trained using SGD optimizer with a learning rate of 0.1 and MultiStepLR scheduler for 200 epochs. EfficientFormer utilized AdamW optimizer with a learning rate of 0.001 for 200 epochs, while ViT was trained with a learning rate of 0.0001. Both transformer-based models used cosine annealing scheduler. All pre-training experiments maintained a 128 batch and were implemented using the Timm library. During pre-training, all quantization operators use INT8, including the baseline, at the initial 25 epochs to ensure a stable start of training. After the initial phase, the precision is modulated to the target bit-width, and training continues.

\begin{table}[!t]
\centering
\small
\captionsetup{font={small}}
\renewcommand{\arraystretch}{1.1}
\begin{tabular*}{\columnwidth}{@{\extracolsep{\fill}}l|ccc@{}}
\toprule
\multicolumn{1}{c}{\textbf{Method}}             & \textbf{Memory}       & \textbf{Acceleration}     & \textbf{Accuracy} \\ \midrule
HOT                                             & 17.48                 & 2.3$\times$               & 93.2           \\
HOT + \textbf{ABC}                             & \textbf{\uline{3.8}}  & 2.3$\times$               & 93.2             \\
HOT + \textbf{ABC} + \textbf{LQS}              & 3.8                   & \textbf{\uline{2.6$\times$}}              & 92.99             \\ \bottomrule \bottomrule 
\end{tabular*}
\caption{Results of incremental ablation study for ViT-B fine-tuning on CIFAR100. Memory represent theoretical calculations, and Acceleration shows the averaged acceleration rates across all layers of ViT-B.}
\label{tab:Incremental ablation}
\vspace{-4mm}
\end{table}

\section{Ablation Study}
To investigate the impact of various components and hyperparameters of HOT on model performance, we conducted comprehensive ablation studies.

\subsection{Incremental evaluation on each component}
\label{sec:Appendix_incremental_ablation}

We analyzed the effects of ABC and LQS in the $g_w$ path of HOT on model accuracy. The experiments were performed by fine-tuning a ViT-B model on the CIFAR100 dataset with 50 epochs, measuring accuracy, memory consumption, and computational speed while incrementally applying both techniques. Here, Memory refers to theoretical calculations, while Acceleration represents the average GPU acceleration across ViT layers. Note that 'HOT' in this experiment refers to the baseline methodology without ABC and LQS.

The results in Table \ref{tab:Incremental ablation} demonstrate that the combination of ABC and LQS effectively optimizes model performance. Specifically, implementing ABC resulted in a significant reduction in memory usage from 17.48GB to 3.8GB, approximately a 79\% decrease. When LQS was additionally applied, the computational speed improved from 2.3$\times$ to 2.6$\times$ while maintaining the reduced memory footprint. Notably, despite these substantial efficiency improvements, the accuracy degradation was limited to merely 0.5\%.

\subsection{Rank selection of HLA}
\label{sec:Appendix_ablation_rank_selection}

We investigated the impact of HLA rank selection on model performance. The experiments were conducted by pre-training EfficientFormer-L1 on the CIFAR100 dataset for 200 epochs, progressively decreasing the number of ranks in powers of two while monitoring performance changes. 

As referred to \cref{tab:HLA_rank}, utilizing 8 ranks in HLA provides optimal accuracy relative to rank reduction. Eight low-frequency vectors appear to be sufficient to represent the spatial information of gradient features. However, further reduction below 4 ranks leads to gradual deterioration, with a particularly sharp decline observed at 2 ranks. Based on these results, we determined that the optimal number of ranks for HLA in HOT implementation is 8.

\subsection{LoRA application method}
\label{sec:Appendix_LoRA_ablation}

\begin{table}[!t]
\centering
\captionsetup{font={small}}
\renewcommand{\arraystretch}{1.1}
    \begin{tabular*}{\columnwidth}{@{\extracolsep{\fill}}ccc@{}}
    \toprule
    \textbf{Selected vectors $r$} & \textbf{Computation Cost} & \textbf{Accuracy} \\ \midrule
    16 (Full rank)         & 1647.48  & 76.35    \\
    8         & 1383.54   & \textbf{\uline{76.25}}    \\
    4         & 1251.56  & 73.09    \\
    2         & 1185.58  & 68.46    \\
    1         & 1152.59   & 47.28    \\ \bottomrule \bottomrule
    \end{tabular*}
\caption{Ablation study of varying the number of low-pass vectors ($r$) in HLA during pre-training of EfficientFormer-L1 on CIFAR100. The optimal $r$ related with computation cost (Gbops) of backward pass is 8.}
\label{tab:HLA_rank}
\vspace{-4mm}
\end{table}

\begin{table}[!t]
\centering
\captionsetup{font={small}}
\setlength{\tabcolsep}{0.5pt}
\renewcommand{\arraystretch}{1.1}
\begin{tabular*}{\columnwidth}{@{\extracolsep{\fill}}ccc@{}}
\toprule
\textbf{\makecell{HOT on \\ Frozen weight}} & \textbf{\makecell{HOT on \\ Decomposed weight}} & \textbf{Accuracy} \\ \midrule
\textcolor{red}{\ding{55}}                      & \textcolor{red}{\ding{55}}                          & 92.61          \\
\textcolor{red}{\ding{55}}                      & \textcolor[RGB]{0,150,0}{\checkmark}                          & 57.96         \\
\textcolor[RGB]{0,150,0}{\checkmark}            & \textcolor{red}{\ding{55}}                          & \textbf{\uline{92.51}}         \\
\textcolor[RGB]{0,150,0}{\checkmark}            & \textcolor[RGB]{0,150,0}{\checkmark}                           & 58.68         \\ \bottomrule \bottomrule
\end{tabular*}
\caption{Experimental results HOT-LoRA combination during fine-tuning ViT-B on CIFAR100. HOT is applied to different combinations of LoRA weight types (Frozen, Decomposed).}
\label{tab:LoRA ablation}
\vspace{-3mm}
\end{table}

\begin{table*}[!t]
    \centering
    \setlength{\tabcolsep}{10pt}
    \renewcommand{\arraystretch}{1.2}
    \captionsetup{font={small}}
    \vspace{-0.1cm}

    \scalebox{0.85}{
        \begin{tabular}{cc|ccccc@{}}
            \toprule
            \textbf{Dataset} & \textbf{Model} & \textbf{FP} & \textbf{INT4} & \textbf{LUQ}~\cite{luq} & \textbf{LBP-WHT}~\cite{lbp-wht} & \textbf{HOT} \\ \midrule
            \multirow{5}*{CIFAR10~\cite{cifar}}   
                       & ResNet-18~\cite{resnet}                  & 95.23 & 93.1 & 94.73 & 93.03 & \textbf{\uline{94.77}}  \\
                       & ResNet-34~\cite{resnet}                  & 95.23 & 93.57 & \textbf{94.74} & 93.59 & 93.83  \\
                       & ResNet-50~\cite{resnet}                  & 94.98 & 90.65 & \textbf{93.62} & 92.12 & 92.85  \\
                       & EfficientFormer-L1~\cite{effformer}         & 95.03 & 92.9 & \textbf{94.1} & 91.07 & 94.01  \\
                       & EfficientFormer-L3~\cite{effformer}         & 95.18 & 93.8 & 94.63 & 91.18 & \textbf{\uline{95.01}}  \\
                       \midrule
            \multirow{5}*{CIFAR100~\cite{cifar}}   
                       & ResNet-18~\cite{resnet}                  & 75.66 & 75.22 & \textbf{75.63} & 72.26 & 75.53  \\
                       & ResNet-34~\cite{resnet}                  & 76.75 & 72.87 & 76.26 & 75.36 & \textbf{\uline{76.95}}  \\
                       & ResNet-50~\cite{resnet}                  & 76.46 & 61.28 & NaN & 69.24 & \textbf{\uline{76.06}}  \\
                       & EfficientFormer-L1~\cite{effformer}         & 76.65 & 64.15 & 75.79 & 73.69 & \textbf{\uline{76.06}}  \\
                       & EfficientFormer-L3~\cite{effformer}         & 77.26 & 66.93 & \textbf{76.19} & 63.04 & \textbf{\uline{76.19}}  \\
                       \midrule
            \multirow{6}*{ImageNet-100~\cite{imagenet}}  
                       & ResNet-18~\cite{resnet}                  & 86.77 & NaN & 82.77 & 82.31 & \textbf{\uline{86.26}}  \\
                       & ResNet-34~\cite{resnet}                  & 86.87 & NaN & 82.6 & 83.31 & \textbf{\uline{86.7}}  \\
                       & ResNet-50~\cite{resnet}                  & 85.51 & NaN & 81.45 & 69.98 & \textbf{\uline{85.2}}  \\
                       & EfficientFormer-L1~\cite{effformer}         & 83.38 & 81.3 & 82.46 & 77.52 & \textbf{\uline{83.05}}  \\
                       & EfficientFormer-L3~\cite{effformer}         & 83.3 & 78.09 & \textbf{83.13} & 78.1 & 83.01  \\
                       & ViT-B~\cite{vit}                   & 77.87 & NaN & 75.97 & 54.25 & \textbf{\uline{77.31}}  \\
                       \midrule
            \multirow{1}*{ImageNet-1k~\cite{imagenet}}   
                       & ViT-B~\cite{vit}                   & 70.01 & NaN & 67.1 & NaN & \textbf{\uline{69.4}}  \\
            \bottomrule \bottomrule 
        \end{tabular}
    }
\caption{Accuracy results for pre-training tasks. In ImageNet training, while other efficiency methods either show substantial degradation (LUQ~\cite{luq}) or fail to train (LBP-WHT~\cite{lbp-wht}), HOT achieves performance nearly equivalent to FP.}
\label{tab:pretraining-extension}
\end{table*}

To determine the optimal strategy for combining HOT with LoRA, we conducted ablation study focusing on frozen weights and decomposed weights. The experiments were conducted under the same conditions as the \cref{sec:Appendix_incremental_ablation}, analyzing four different configurations based on combinations of HOT to these two weight types.

The results presented in \cref{tab:LoRA ablation} validate the effectiveness of our proposed HOT-LoRA integration. The configuration applying HOT exclusively to frozen weights achieved the highest accuracy of 92.51\% among all tested combinations. In contrast, configurations that applied HOT solely to decomposed or to both decomposed and frozen weights shows significant performance degradation. These findings experimentally demonstrate that direct training of decomposed weights plays a crucial role in model performance.

\section{Overhead Calculation}
\label{sec:Appendix_overhead_calculation}

To analyze the computational overhead of HOT, we adopt a standardized notation for layer dimensions. Each layer is represented as a tuple (L, O, I), where L denotes the spatial dimension size, O represents the output channel size, and I indicates the input channel size. For instance, in a layer denoted as (128, 64, 256), the spatial dimension is 128, with 64 output channels and 256 input channels.

The additional transformation, reshaping, and quantization/dequantization processes introduce some overhead to HOT compared to the vanila BP. However, by leveraging low-precision arithmetic, we can achieve practical performance benefits. In this section, we present three tables to explain the computational benefits of our approach in detail.

The added overhead of HOT can be negligible, as shown in Table \ref{tab:overhead_breakdown}, especially when $\log n$ is sufficiently small relative to other dimensions. In our work, we use $n = 16$ for applying order-4 block-diagonal HT, making this condition valid. For example, in the 'stages.3.fc2' (49, 448, 1792) layer of EfficientFormer-L1, vanilla BP requires 137.3 MFlops, whereas our method only requires 11.5 MFlops. Although HOT incurs some additional overhead, it can achieve significant computational cost reduction through efficient low-precision arithmetic.

\section{Pre-training result on various models}
\label{sec:Appendix_pretraining}

In this section, we present additional experimental results on pre-training task, evaluated across various architectures and datasets. As shown in Table~\ref{tab:pretraining-extension}, HOT consistently outperforms naive INT4 training and previous methods in almost all scenarios, following trend of fine-tuning result.

HOT surpasses LBP-WHT~\cite{lbp-wht} in all cases, highlighting the importance of proper optimization of each gradient paths. LUQ~\cite{luq} achieves competitive performance on the CIFAR10 dataset~\cite{cifar}, but it occasionally fails to train or exhibits significantly degraded results on more complex datasets, such as CIFAR100~\cite{cifar} and ImageNet~\cite{imagenet}. 

Notably, HOT is the only approach that provides stable training while achieving performance comparable to FP training. The extensive pre-training results demonstrate that HOT is a robust and comprehensive solution, applicable not only to fine-tuning but also to pre-trainig tasks.

\section{Details of CUDA kernel}
\label{sec:Appendix_CUDA_kernel}

\textbf{Implementation: }This section details the implementation of CUDA kernels in HOT. The CUDA kernels consist of five core modules: Hadamard transformation(HT), Hadamard Low-rank Approximation(HLA), Quantization, INT4 and INT8 matrix multiplication, and Dequantization. 

The HT and HLA kernel, which inherently has an efficient computational complexity of $O(nlogn)$ with FWHT algorithm, is optimized to maximize computational efficiency by extensively utilizing shared memory of GPU to minimize memory access latency. FWHT is employed not only for HT but also for HLA, where vectors corresponding to low-pass vector indices selected based on the $LP_{L1}$ criterion are extracted from FWHT outputs and concatenated into a single tensor. 

The Quantization module implements pseudo-stochastic quantization\cite{wang2022niti}, ensuring unbiased estimation while minimizing quantization overhead. 

Both INT4 and INT8 matrix multiplication are implemented using the NVIDIA CUTLASS framework. Unlike CUBLAS, which is closed-source, CUTLASS is open-source and provides extensive customization options for matrix operation configurations, enabling the implementation of optimized matrix multiplication with high throughput. Specifically, to address PyTorch's lack of native INT4 data type support, we efficiently compressed tensors by packing two INT4 values adjacently within an INT8. 

Finally, for the Dequantization stage, which requires matrix multiplication in FP32 format, we utilized NVIDIA CUBLAS through PyTorch's default matrix multiplication implementation.

\begin{figure}[t]
     \centering
         \includegraphics[width=1.0\columnwidth]{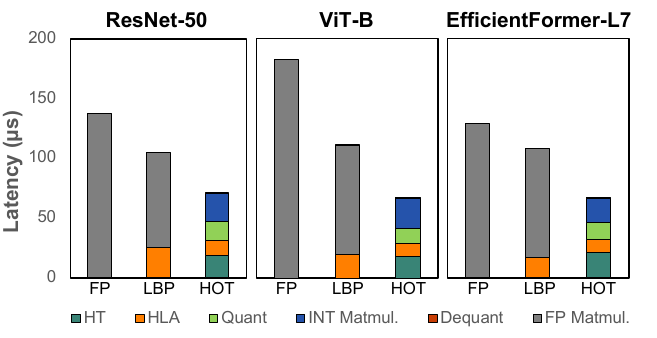}
     \caption{Comparison of kernel latency ($\mu$s) between FP32, LBP-WHT, and HOT for representative layers of ResNet-50, ViT-B, and EfficientFormer-L7 models. The selected representative layers can be found in the 'Latency breakdown' section of \cref{sec:Appendix_CUDA_kernel}.}
     \label{fig:latency_breakdown}
     \vspace{-4mm}
\end{figure}

\textbf{Latency breakdown: }In this section, we conduct a detailed latency analysis of five core modules in CUDA kernels to understand the specific mechanisms of computational acceleration. \cref{fig:latency_breakdown} presents a comparative analysis of latencies across FP operations, LBP-WHT, and HOT kernels for representative layers showing average acceleration in ResNet-50, ViT-B, and EfficientFormer-L7 architectures. Specifically, HOT achieved 1.9$\times$ acceleration in ResNet-50's 'layer4.conv2' (49, 512, 4608), 2.6$\times$ acceleration in ViT-B's 'qkv' (197, 2304, 768), and 1.9$\times$ acceleration in EfficientFormer-L7's 'stages.1.fc1' (784, 768, 192).

The experimental results demonstrate that integer matrix multiplication significantly
reduced the latency compared to FP32 across all target models. A noteworthy case is observed in the ViT-B model, where FP32 matrix multiplication shows 182$\mu s$, while integer matrix multiplication consumes only 25$\mu s$. This substantial performance improvement can be attributed to the synergistic effect of low-precision integer matrix multiplication combined with tensor size reduction through HLA.

Meanwhile, the HT and HLA modules in ViT-B show latencies of 18$\mu s$ and 11$\mu s$ respectively, resulting in approximately 16\% computational overhead compared to FP operations. This figure exceeds the theoretical overhead prediction of 7\% calculated in \cref{sec:Appendix_overhead_calculation}. This discrepancy can be attributed to additional operations required in practical implementation beyond FWHT algorithm operations, including transpose, reshape, and contiguous operations. Consequently, we anticipate potential further latency reductions through the optimization techniques such as kernel fusing.

\section{Various case of output gradient tensor}
\label{sec:Appendix_gradient_tensor_case}

\begin{figure*}[!t]
    \centering
    \includegraphics[width=\linewidth]{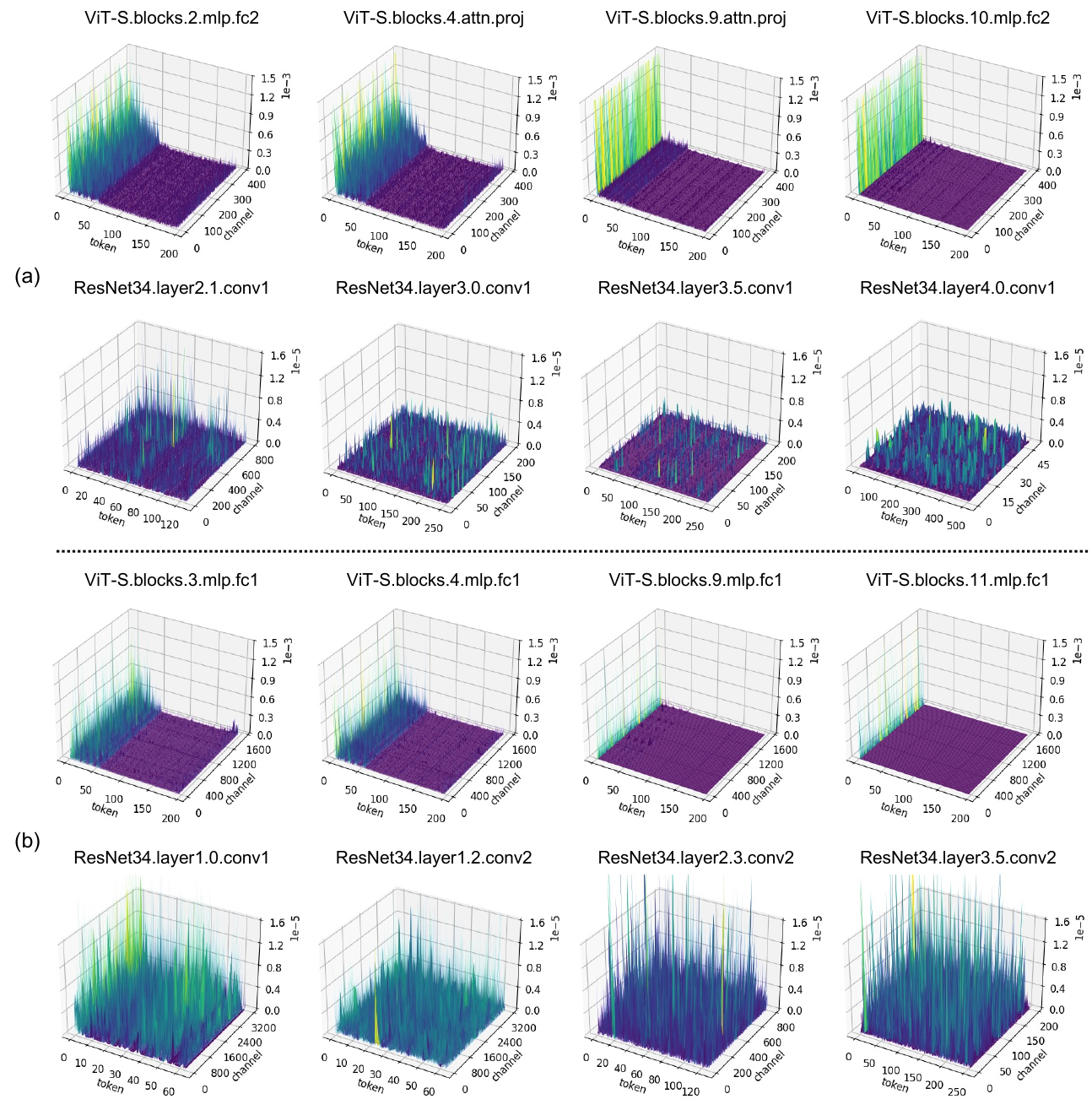} 
    \caption{The illustration for output gradient of (a) Per-token quantization friendly case and (b) Per-tensor quantization friendly case.}
    \label{fig:Appendix_Per_Tensor_or_Per_Token}
    \vspace{-4mm}
\end{figure*}

We visualized two distinctive outlier patterns of output gradient $g_y$ that form the theoretical baseline of LQS, extending our analysis to various layers within ViT-S and ResNet34. \cref{fig:Appendix_Per_Tensor_or_Per_Token} presents 3D visualizations from ImageNet-1k training with LQS, categorized layers into (a) per-token quantization friendly and (b) per-tensor quantization friendly case.

The analysis revealed distinct pattern across models. In ViT-S, fc2 layers and attention projection layers consistently exhibited token-wise gradient outliers, demonstrating it is suited for per-token quantization. Similarly, conv1 layers across different stages in ResNet-34 displayed non-zero gradient value at the token level, indicating the effectiveness of per-token quantization for these layers.

Conversely, some layers exhibited contrasting characteristics. In the case of ViT-S's fc1 layers, there was a significant reduction in the magnitude of token-level non-zero gradients compared to fc2 layers, with notably sparse occurrence patterns, suggesting limited effectiveness of per-token quantization. Certain conv1 and conv2 layers in ResNet-34 showed large non-zero gradients at irregular positions independent of token locations, indicating limited advantages in terms of quantization error when applying per-token quantization.

These analysis results empirically demonstrate the necessity for different quantization strategies based on layer characteristics, suggesting that LQS represents an optimized approach that considers characteristics of each layer.

\begin{table}[!t] 
    \setlength{\tabcolsep}{0.3em}
    \small
    \captionsetup{font={small}}
    \centering
    \setlength{\tabcolsep}{1pt}
    \vspace{-1mm}
    \begin{tabular*}{\columnwidth}{@{\extracolsep{\fill}}cr@{}}
    \toprule
    \multicolumn{1}{c}{\textbf{Name}} & \multicolumn{1}{c}{\textbf{FLOPs}} \\ \midrule
    Vanilla BP                        & $4LIO$                            \\ \midrule
    $g_x$                      & $2LOlogn + 2IOlogn +2LO + 2IO$                       \\ \midrule          
    $g_w$                      & $2LIlogn + 2LOlogn +2I(L*\frac{r}{n}) + 2O(L*\frac{r}{n})$                            \\ \midrule
    Dequant                           & $2IO + 2LI$                             \\ \bottomrule
    \end{tabular*}
    \caption{The additional FLOPs induced by optimization path.}
    \label{tab:overhead_breakdown}
    \vspace{-1mm}
\end{table}

\clearpage

 \clearpage
{
    \small
    \bibliographystyle{ieeenat_fullname}
    \bibliography{main.bbl}
}


\end{document}